\documentclass{article}

\PassOptionsToPackage{numbers, compress}{natbib}

\usepackage[preprint]{neurips_2023}

\usepackage[utf8]{inputenc} 
\usepackage[T1]{fontenc}    
\usepackage{natbib}
\usepackage{hyperref}       
\usepackage{url}            
\usepackage{booktabs}       
\usepackage{amsfonts}       
\usepackage{nicefrac}       
\usepackage{microtype}      
\usepackage{xcolor}         
\usepackage{graphicx}
\usepackage{tabularx}
\usepackage{subcaption}
\usepackage{amsmath}
\usepackage{authblk}

\newcommand{\SHREYANK}[1]{\textcolor{red}{[Shreyank: #1]}}
\newcommand{\ANURAG}[1]{\textcolor{blue}{[Anurag: #1]}}
\newcommand{\JON}[1]{\textcolor{green}{[Jon: #1]}}

\renewcommand{\SHREYANK}[1]{}
\renewcommand{\ANURAG}[1]{}
\renewcommand{\JON}[1]{}

\title{Optimizing ViViT Training: Time and Memory Reduction for Action Recognition}

\author[1]{\textbf{Shreyank N Gowda}\thanks{Work done as a Student Researcher at Google.}}
\author[2]{\textbf{Anurag Arnab}}
\author[2]{\textbf{Jonathan Huang}}
\affil[1]{University of Edinburgh}
\affil[2]{Google Research}

\begin{document}

\maketitle

\begin{abstract}
In this paper, we address the challenges posed by the substantial training time and memory consumption associated with video transformers, focusing on the ViViT (Video Vision Transformer) model, in particular the Factorised Encoder version, as our baseline for action recognition tasks. The factorised encoder variant follows the late-fusion approach that is adopted by many state of the art approaches. 
Despite standing out for its favorable speed/accuracy tradeoffs among the different variants of ViViT, its considerable training time and memory requirements still pose a significant barrier to entry.
Our method is designed to lower this barrier and is based on the idea of freezing the spatial transformer during training. This leads to a low accuracy model if naively done.  But we show that by (1) appropriately initializing the temporal transformer (a module responsible for processing temporal information) (2) introducing a compact adapter model connecting frozen spatial
representations ((a module that selectively focuses on regions of the input image) to the temporal transformer, we can enjoy the benefits of freezing the spatial transformer without sacrificing accuracy.
Through extensive experimentation over 6 benchmarks, we demonstrate that our proposed training strategy  significantly reduces training costs (by $\sim 50\%$) and memory consumption while maintaining or slightly improving performance by up to 1.79\% compared to the baseline model. 
Our approach additionally unlocks the capability to utilize larger image transformer models as our spatial transformer and access more frames with the same memory consumption. The advancements made in this work have the potential to propel research in the video understanding domain and provide valuable insights for researchers and practitioners with limited resources, paving the way for more efficient and scalable alternatives in the action recognition field.

\end{abstract}

\section{Introduction}
\label{intro}

\begin{figure}
  \centering
  \includegraphics[width=0.85\linewidth]{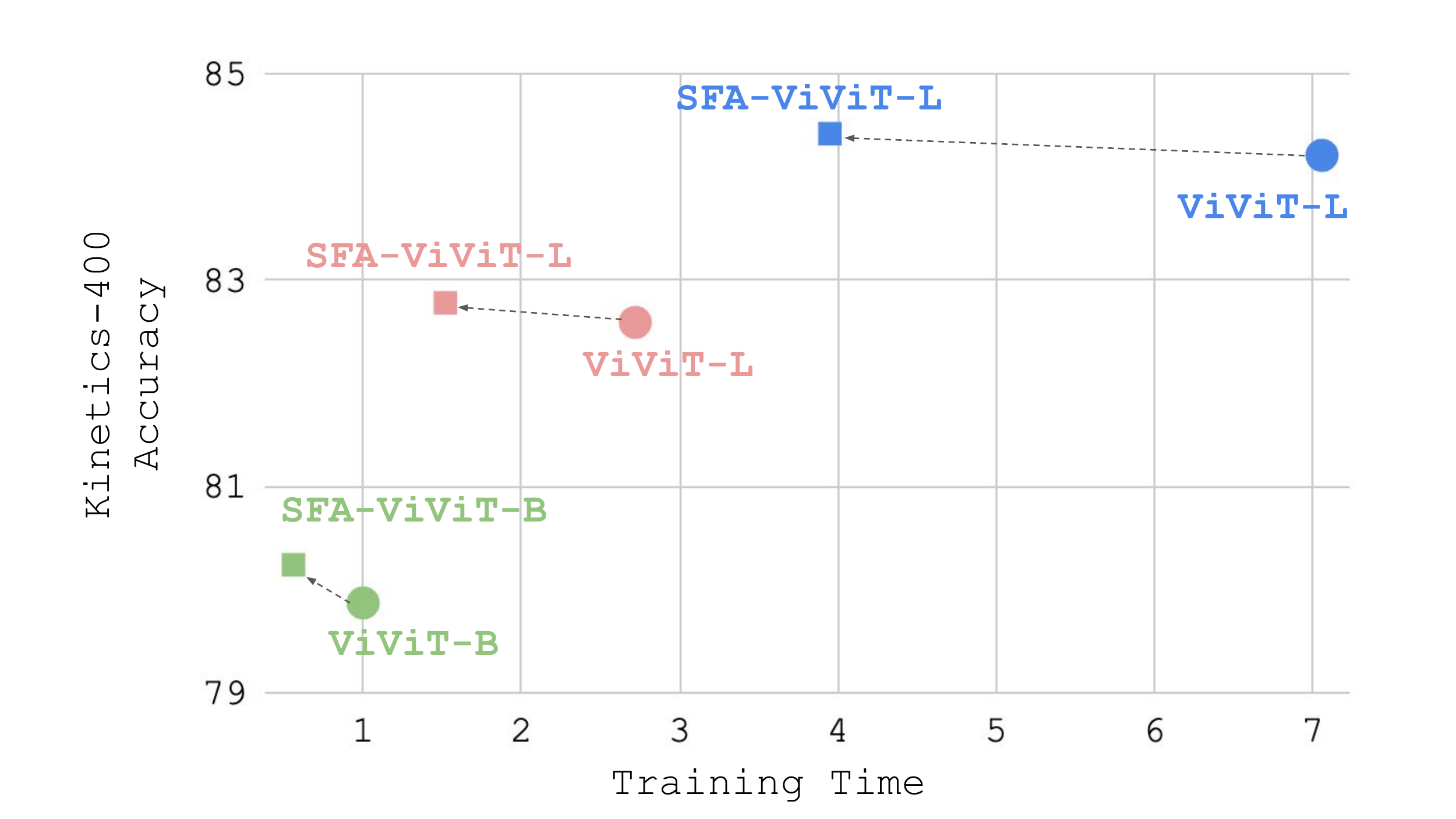}
  \caption{Comparison of our initialization method vs conventional training of ViViT. Training time is scaled relative
  to setting ViViT-B training time to `1 unit' (7.93 hours).
  We see clear time saving using our initialization scheme and for larger models, the training time saved is much larger. 
  }
  \label{fig:teaser}
\end{figure}

Action recognition 
focuses on understanding and identifying  actions in video sequences with applications in surveillance, human-computer interaction, and video content analysis. The field has advanced significantly due to 
large-scale annotated datasets~\cite{i3d}
and a shift from hand-crafted features~\cite{laptev2005space,wang2013dense} to deep learning models like convolutional networks (CNNs)~\cite{simonyan2014two,tran2015learning,i3d,gowda2017human}.
Recently, transformers have revolutionized computer vision by offering an alternative to traditional CNNs, 
leading to the development of many new state-of-the-art architectures~\cite{vit, detr, swin}. Moreover, the flexibility of transformers have inspired researchers to adapt these models to more complex problems, including video understanding and action recognition~\cite{vivit,videoswin}.

Transformers, however, are notoriously expensive, and Video Transformer-based architectures~\cite{timesformer,vivit,videoswin}, which integrate information across space
and time are even more so. And memory consumption and training times become even more significant when working with large-scale video datasets with long sequences \cite{i3d}.
These high computational costs present a particular 
challenge for researchers with limited resources, especially those from universities and smaller companies.
The goal of our work is therefore to cut the cost of training: we want to train transformer-based video models with fewer resources or use larger model variants and handle more frames with the same resources.


We have chosen the ViViT model \cite{vivit} as our baseline upon which to improve.  Specifically, we focus on the ``factorized encoder'' variant of ViViT which has separate spatial and temporal transformer stages, where the spatial transformer is responsible for extracting features from individual frames, while the temporal transformer processes the temporal dynamics across frames. We choose this factorized encoder design because it is more efficient compared to, e.g., the variant of ViViT using all-to-all spatiotemporal attention, while still achieving high accuracies and has thus been adopted as the building block for recent state-of-the-art architectures on various tasks~\cite{mtv,chen2022mm,wanggit,xiong2022m,yang2023vid2seq,gritsenko2023end}. 

To address the challenge of reducing training time and memory usage without compromising the  sophistication and accuracy of  the  original  model, our approach is based on the simple idea of freezing the spatial backbone.  Freezing the spatial backbone has many advantages: by not backpropagating through this transformer, training is faster and requires less memory (allowing for the model to handle more frames).  We also inherit the benefits of pretraining the spatial transformer on a large dataset (such as JFT~\cite{jft}).  Naively implemented, however, we show that this approach falls very short in accuracy.  Instead, with a few simple (but important) tweaks to the above idea, we propose a method that has the same advantages of freezing the spatial transformer, but does not compromise on accuracy.

Our method proceeds in two stages. In the first stage we pretrain a cheap version of the model using fewer frames, e.g., 8 frames as opposed to, e.g., 32 frames. 
In the second stage we fine tune this model with more frames, which is more expensive but in this stage we freeze the spatial encoder and introduce a compact ``adapter'' model
connecting frozen spatial representations to the temporal transformer, negating the need for end-to-end training of the spatial transformer. Critically this includes pre-training the temporal transformer (by initializing from stage 1) which is often overlooked in current video models which typically initialize this component from scratch. However  our experiments show that this step is critical if we wish to not sacrifice performance.


Drawing parallels with curriculum learning~\cite{bengio2009curriculum},
our methodology can also be viewed as progressively training 
on tasks of increasing complexity, beginning with a ViViT model pre-trained on 8 frames — our ``easy examples''. As we progress, the model effectively handles larger frame counts up to 128 frames - our ``difficult examples''. This approach not only sustains the intricacy of the original model but also significantly reduces resource demands. Thus our approach enables entities with limited resources to emulate high-performance models using affordable GPUs.

With our training recipe, we match or slightly outperform conventional training of ViViT at roughly half the cost as seen in Figure~\ref{fig:teaser}. A notable benefit of our training recipe, is its ability to process up to 80 frames on typical university-grade GPUs, a significant leap from the previous capacity of 16 frames. This expansion in processing power broadens the range of video data manageable under resource-constrained settings. As we elaborate in Section 4.8, our research underscores the potential to democratize access to advanced video transformer models. Another notable benefit is the model's ability to now use even larger models as the spatial transformer, we introduce ViViT-g as seen in Section~\ref{sec:vivit_g}. This accessibility paves the way for future video action recognition research, irrespective of resource constraints. Hereafter, we refer to our version of ViViT as SFA-ViViT, where SFA denotes '\textbf{S}patial \textbf{F}rozen and \textbf{A}dapter Initialized'.

\SHREYANK{Updated the previous paragraph to include examples and reference curriculum learning rather than counter distillation. Not sure about the model/training name.}

\section{Related Work}
\label{background}

\paragraph{Transformers for Videos} Action recognition is a key research area in computer vision, addressed by many traditional~\cite{laptev2005space,wang2013dense} and CNN based approaches~\cite{i3d,smart,ji20123d,lin2019tsm,simonyan2014two,tran2015learning,tsn,l2a} aided by the release of large-scale datasets~\cite{i3d,kuehne2011hmdb,soomro2012ucf101}. Since we focus on transformer based architectures, a thorough review of earlier methods are out of this scope. More recently, the transformer architecture, initially developed for NLP tasks \cite{vaswani2017attention}, has been adapted for video understanding and action recognition tasks, leading to state-of-the-art models such as TimeSformer \cite{timesformer}, ViViT \cite{vivit}, VideoSwin \cite{videoswin}, and Uniformer \cite{li2022uniformer} 
These transformer-based models leverage self-attention mechanisms to capture complex spatiotemporal patterns in action recognition tasks. 
TimeSformer \cite{timesformer} is one of the first transformer-based models for video understanding, adapting
the transformer architecture to video by treating it as a sequence of flattened image patches. ViViT \cite{vivit} integrates spatial and temporal transformers to efficiently capture spatiotemporal information in video sequences. VideoSwin \cite{videoswin} is a hierarchical transformer  that applies local windowing for efficiency, 
enabling the model to handle longer video sequences. VideoBERT \cite{sun2019videobert} is a transformer model that learns joint representations of video and language in self-supervised manner, which can be fine-tuned for various video understanding tasks, including action recognition. 
More recently, Uniformer \cite{li2022uniformer} integrates 3D convolution and spatiotemporal self-attention, 
MTV~\cite{mtv} proposes a multi-view transformer model using distinct encoders for each video ``view'', improving accuracy as the number of views increases.
The Multiscale Vision Transformers (MViT)~\cite{mvit} model streamlines computation and memory usage by operating at different resolutions, focusing on high-level features at lower resolutions and low-level details at higher ones, effectively leveraging both spatial and temporal information in visual tasks. TubeViT~\cite{tubevit} introduces a method of sparsely sampling different-sized 3D segments from videos, facilitating efficient joint image and video learning, and allowing the adaptation of larger models to videos with less computational resources. Typically these models have FLOPs in the range of TFLOPs and training times that last more than days on the largest of GPUs/TPUs available, making them infeasible to train or use in 
lower resourced settings such as academia. It is critical that we find a way to train 
these models with limited resources while maintaining their performance. To this end, we focus on the factorised encoder version of ViViT as the late-fusion approach followed is used as a foundation for state-of-the-art approaches of various tasks~\cite{mtv,chen2022mm,wanggit,xiong2022m,yang2023vid2seq,gritsenko2023end} and hence believe that the initialization scheme proposed can be used for future methods working on similar architectures.
\ANURAG{Want to highlight more about the computational cost of these models, and what people have proposed to deal with it. Other relevant papers could be MTV, MViT, TubeViT}\SHREYANK{Do you mean mention GFLOPS, params etc to compare? Added MTV, MVIT and TubeVIT}
\ANURAG{I think you need to mention here why this paper now focuses on ViViT instead of the other alternatives.}\ANURAG{An obvious reviewer question is why did the authors not experiment on the latest architectures}
\ANURAG{One possible idea could be to mention that the ViViT-FE / late-fusion approach is used in a lot of other approaches, and so its of general interest? Also, MTV builds upon ViViT-FE}
\ANURAG{And yes, mention a bit more about the cost of these models being a problem. ie to highlight the problems in the field which this paper is addressing}
\SHREYANK{Is this better now? Just added a line about tflops and over 24 hours training time and that is infeasible for people with low resources.}

\paragraph{Efficient Transformers in Videos}
\ANURAG{For each of the papers, say what they are actually efficient in: inference time, training time, FLOPs, memory usage?}

Efficiency is a nuanced topic~\cite{efficiencymisnomer}, as there are multiple cost indicators of efficiency (for example, GFLOPs, inference time, training time, memory usage), and models which improve efficiency in one dimension, are not necessarily better in other dimensions~\cite{efficiencymisnomer}.
TokenLearner~\cite{ryoo2021tokenlearner} proposes a method that adaptively learns tokens for efficient image and video understanding tasks, enabling effective modeling of pairwise attention over longer temporal horizons or spatial content. TokenLearner reduces the GFLOPs required by ViViT by about half, but does not significantly change the training time or the inference time of ViViT.
Spatial Temporal Token Selection (STTS) \cite{stts} proposes a dynamic token selection framework for spatial and temporal dimensions that ranks token importance using a lightweight scorer network, selecting top-scoring tokens for downstream evaluation in an end-to-end training process. STTS again reduces the GFLOPs, but the training time and inference time do not change significantly. TokShift \cite{tokshift}, a zero-parameter, zero-FLOPs operator that models temporal relations in transformer encoders by temporally shifting partial token features across adjacent frames but again requires the same training time as the original model. By densely integrating TokShift into a plain 2D vision transformer, a computationally efficient, convolution-free video transformer is created for video understanding. Most similar to our work is the ST-Adapter \cite{adapter}, that utilizes built-in spatio-temporal reasoning in a compact design, allowing pre-trained image models to reason about dynamic video content with a small per-task parameter cost, surpassing existing methods in both parameter-efficiency and performance. However, it does not change FLOPs or inference time at all. Unlike ST-Adapter, we use a spatial only adapter which we show is enough to reproduce the performance of the baseline model at close to half the training time. In particular, our proposed method improves the training time and training memory usage, addressing the key problem of researchers and practitioners being able to train video models. It does not, however, change the inference time compared to a standard ViViT model.
\ANURAG{Doesn't necessarily fit in this place, but something like: ``Efficiency is a nuanced topic~\cite{efficiencymisnomer}, as there are multiple cost indicators of efficiency (for example, GFLOPs, inference time, training time, memory usage), and models which improve efficiency in one dimension, are not necessarily better in other dimensions~\cite{efficiencymisnomer}. In particular, our proposed method improves the training time and training memory usage, addressing the key problem of researchers and practitioners being able to train video models.
It does not, however, change the inference time compared to a standard ViViT model''}
We consider overall train time for the same hyperparameters and use the same hardware for a direct comparison. We consider efficiency in this paper as the time saved in the overall training of the model. 

\ANURAG{Also need to be careful when talking about efficiency: i.e efficiency in terms of the number of parameters, FLOPs, training time, inference time? They are different and doing well in one does not mean doing well in the other. For example, ST-Adapter is parameter-efficienct, but does not change FLOPs or inference time at all. Also check The Efficiency Misnomer for some more details}\SHREYANK{Added something for this, let me know what you think.}
\ANURAG{My main point is you need to specify efficient in terms of what. For example, TokenLearner is efficient in terms of GFLOPs. But if you look at the actual training / inference time, it does not change much. Similarly, ST-Adapater, is efficient in terms of parameters, but the training and inference time, also GFLOPs are identical to a standard ViViT. Papers typically report what dimensions they are efficient in terms in, but not in terms of what they are not.}
\ANURAG{And so my main point here is to be clear in terms of what this proposed approach is efficient in. For example, it improves the training time. It does not change the GFLOPs or inference time of the model though.}
\SHREYANK{I've mentioned that now, do I also need to mention that we don't change flops or inference time?}

\section{Methodology}
\label{method}

\subsection{Revisiting ViViT}

The Video Vision Transformer (ViViT) extends the Vision Transformer architecture to handle video data by incorporating spatio-temporal reasoning. The idea behind ViViT is to process video input as a sequence of image patches, combining spatial and temporal information through a series of transformer layers, which include multi-head self-attention, layer normalization, and feed-forward networks. The output is used for video classification.

In the ``vanilla'' variant of ViViT, one extracts
spatio-temporal tokens from a video then forwards all 
tokens
through a transformer encoder which explicitly
models all pairwise interactions between all spatio-temporal tokens.  We build off of the more efficient
``Factorized Encoder'' variant of ViViT whose
architecture consists of two separate transformer encoders,
a \emph{spatial transformer} modeling interactions between tokens from the same temporal index and a \emph{temporal transformer} modeling interactions between tokens
from different temporal indices. Despite having more parameters, it requires fewer floating point operations (FLOPs) than 
vanilla ViViT.
Because the Factorised Encoder variant strikes a
good balance point between accuracy and processing speed,
it has also been adopted as the foundation for other architectures\cite{mtv,chen2022mm,wanggit,xiong2022m,yang2023vid2seq,gritsenko2023end}, reinforcing its utility and robustness.

\ANURAG{Just say that this paper focuses on training time. And that this is a key bottleneck in both research and industry. But the actual model stays the same, so inference time and GFLOPs are unchanged from a standard ViViT.}
\ANURAG{The main point of the Efficiency Misnomer is to be explicit about what dimensions you are ``efficient in'', and if you are claiming to be ``efficient'' in general, then you need to report all the efficiency metrics.}
\SHREYANK{Updated.}

\subsection{Our training strategy}

We concentrate on the factorised encoder variant of ViViT as it is already the most efficient version of the baseline. Henceforth, when we talk about ViViT we refer to this variant of ViViT. 
Consider the ViViT model that contains a spatial transformer with parameters $\theta_{spatial}$ and a temporal transformer with parameters $\theta_{temporal}$:
%
\begin{align}
    X_{spatial} &= T_{spatial}(X_{in}; \theta_{spatial}) \\
    X_{out} &= T_{temporal}(X_{spatial}; \theta_{temporal}).  \nonumber  
\end{align}
In conventional ViViT training, $\theta_{spatial}$ is initialized from an image pre-trained checkpoint such as ImageNet-21k~\cite{ridnik2021imagenet} or JFT~\cite{jft} and the $\theta_{temporal}$ is initialized from scratch. 
During backpropagation, the gradient flows through the entire model. This entails training two sizable transformer models end-to-end, which is a highly resource-intensive process, as the transformer architecture is inherently computationally demanding,
especially with more frames and  larger ViViT variants (e.g., 
ViViT-H).

One approach to reducing training time is to 
freeze the parameters of the spatial transformer
$\theta_{spatial}$.  
By not backpropagating through $\theta_{spatial}$, 
gradient updates are faster and require less memory, allowing us to access more frames without encountering
out-of-memory issues.
But as we show in experiments, the accuracy of the resulting model with frozen $\theta_{spatial}$
is not competitive (in accuracy) with the baseline training approach.

\begin{figure}
  \centering
  \includegraphics[width=0.95\linewidth]{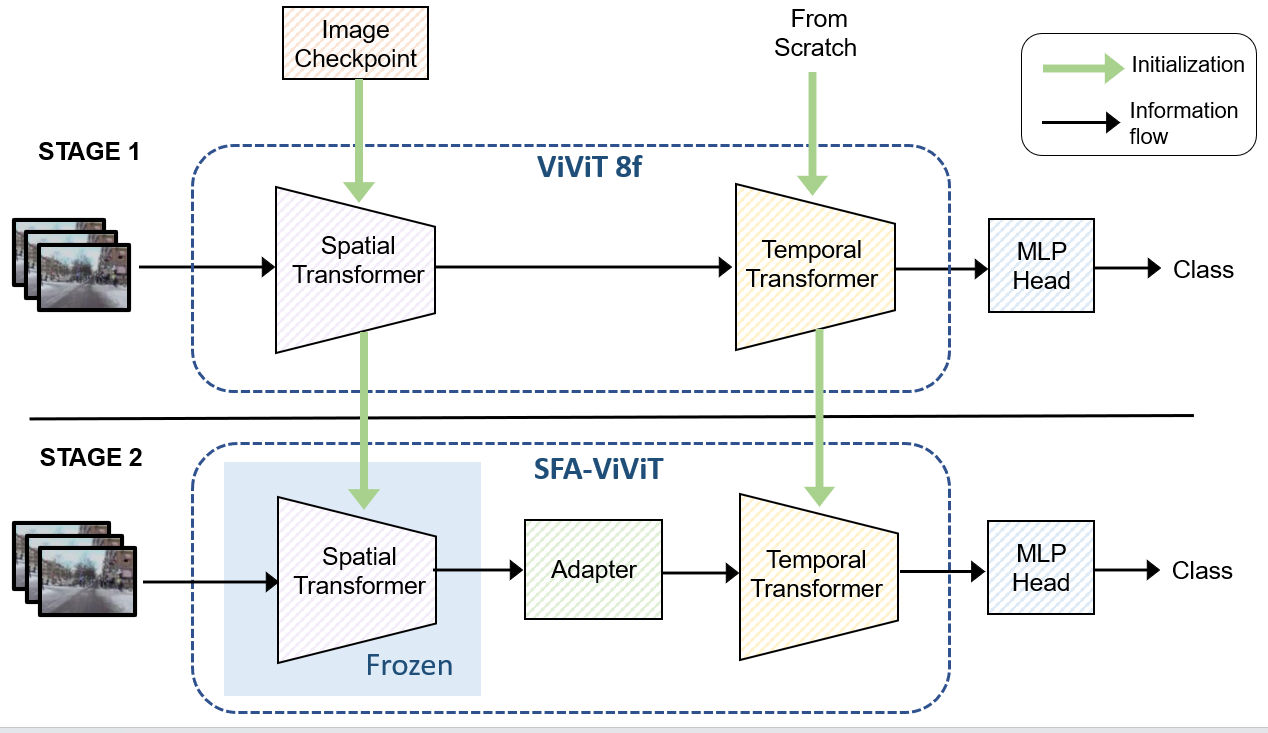}
  \caption{STAGE 1: We first use the full ViViT-FE model on 8 frames by initializing the spatial transformer from an image checkpoint and the temporal transformer from scratch. STAGE 2: We then use this as our checkpoint to initialize the spatial and temporal transformer for models using more frames (such as 32, 64 or 128). We then freeze the spatial transformer and add an adapter model to finetune spatial transformer features. The temporal transformer is finetuned from the same checkpoint. 
  }
  \label{fig:overview}
\end{figure}

We present a two stage approach
(see
Fig.~\ref{fig:overview})
to training ViViT models that inherits 
the same benefits of freezing the spatial transformer, 
while not compromising on model quality. 

\paragraph{Stage 1.}
In Stage 1, we pretrain our ViViT model on a reduced number of frames initializing the spatial
transformer using a pre-trained image 
checkpoint.  We do not freeze the spatial transformer
during this stage, but critically, Stage 1 
serves to also initialize the temporal transformer.

To set the number of frames at this stage, we must balance the goal of efficiency (using fewer frames)
against our finding in experiments that pre-training
on too few frames can lead to suboptimal results.
In our ablations, we identify a sweet spot at 8 frames.

\paragraph{Stage 2.} In Stage 2, 
we fine tune our ViViT model
on the full frame count (e.g. 128 frames) initializing
both spatial and temporal 
transformer parameters learned in Stage 1.
Because this stage is significantly more expensive, in stage 2, we freeze the spatial transformer parameters
$\theta_{spatial}$ and add a lightweight adapter module
with parameters $\theta_{adapter}$  following the spatial transformer:  
\begin{align}
    X_{spatial} &= T_{spatial}(X_{in}; \theta_{spatial}) \nonumber \\
    X_{adapter} &= A_{adapter}(X_{spatial}; \theta_{adapter}) \\
    X_{out} &= T_{temporal}(X_{adapter}; \theta_{temporal})  \nonumber
\end{align}

In this setting, 
by  backpropagating only through the temporal transformer and the lightweight adapter module
(in our experiments, a two layer MLP), we
effectively cut total training time by half.

\ANURAG{Include a figure to describe this process}
\SHREYANK{Not sure what type of figure I should use here. Do you mean a flowchart going from each stage to the next? I've added an example on what I mean.}
\ANURAG{I meant the actual training process. ie Image initialsation $\to$ train full model with 8 frames $\to$ add adaptor, and freeze spatial transforemr $\to$ then train the temporal transformer}
\ANURAG{And then you are claiming that all of these steps still take less time then training ViViT the standard way? You should definitely include training curves showing this.}
\SHREYANK{Added the updated figure.}

The crucial finding here is that the spatial transformer requires only short-term context for initialization (after which it remains frozen), whereas the temporal transformer necessitates long-term context to achieve its optimal performance. Further details and empirical analysis can be found in the next section.

\section{Experimental Analysis}
\label{exp}

Through a series of comprehensive experiments
which we now present,
we investigate the significance of the spatial transformer, examining the impact of pre-training datasets and how larger models affect action recognition performance. We also explore the importance of initializing the temporal transformer by employing various initialization schemes and datasets, assessing whether the number of frames is critical for initializing larger models, initializing full ViViT models, and initializing models on one dataset while fine-tuning on another.

\subsection{Datasets}

We evaluate on all the datasets considered in ~\cite{vivit} 
(specifically, Kinetics-400~\cite{i3d}, Kinetics-600~\cite{k600}, EPIC-Kitchens~\cite{epic}, Something-something v2~\cite{ssv2} and Moments-in-time~\cite{mit})
as well as the Something-Else~\cite{selse} dataset. 
As these datasets are common in the community, we include further details in the supplementary.
\SHREYANK{Added this to save space, but want to check what you guys think.}

\subsection{Implementation Details}

We use Scenic~\cite{dehghani2022scenic} for our implementation. Since we build on ViViT, we directly work on top of the codebase and stick to the default parameters used by ViViT in terms of hyperparameters. Full details of these can be found in supplementary.

Our adapter is a two-layer fully connected network that takes as input the output from the spatial transformer and the output from the adapter is passed as input to the temporal transformer.

The hyper-parameters of the transformer models are set to the standard: the number of heads are 12/16/16/16, number of layers are 12/24/32/40, hidden sizes are 768/1024/1280/1408 and MLP dimensions are 3072/4096/5120/6144 for the base/large/huge/giant versions respectively. The 8-frame ViViT model is trained for 30 epochs. We also experiment with initializing larger models with an 8-frame model trained for 10 epochs. Details of this can be found in the supplementary.

For our hardware, we use 64 v3 TPUs for  all  experiments. However, we also show results using 8 NVIDIA GeForce 2080 Ti (w/12 GB memory). This is a typical setting in a small academic lab.
\JON{details on how long you do 8 frame pretraining?}

\subsection{Ablation Study}
We first address two critical aspects: the significance of fine-tuning the spatial transformer and the importance of initializing the temporal transformer. To do so, we conduct a series of experiments in various scenarios, which are detailed below. Our analysis focuses on the Something-something dataset, utilizing the large version of the ViViT model, referred to as ViViT-L.

\ANURAG{What is ``convential ViVIT training''}
\ANURAG{I think it might be clearer here to introduce some notations, ie the model paramters can be $\theta_{spatial}$, $\theta_{temporal}$, $\theta_{head}$. And then you talk about how these are initialised and trained.}
\ANURAG{In general, I find the following paragraphs unclear}
\SHREYANK{Is it better now? I've updated the para with notations.}
\ANURAG{Yes, its better. You can move some of the notation to the method section.}

We examine four main elements that modify the structure of conventional ViViT training and these are mentioned with indices in Table~\ref{tab:ablation} namely: I. The freezing of the spatial transformer ($\theta_{spatial}$ is initialized and then frozen), II. The freezing of the temporal transformer ($\theta_{temporal}$ is frozen), III. The addition of an adapter (lightweight module with parameters $\theta_{adapter}$),  IV. Next, we initialize the temporal transformer using VideoMAE\cite{videomae}, while keeping the spatial transformer frozen and the adapter incorporated and V. The initialization of the temporal transformer ($\theta_{spatial}$ and $\theta_{temporal}$ are initialized using the 8-frame version of the baseline). 

It is important to note that the VideoMAE training is an extremely expensive process as can be seen in the table.
But combined with the line below it, these two models, which
significantly outperform lines I, II and III, show that 
properly initializing the temporal transformer is the critical
issue at hand.

\ANURAG{I think its clearer to describe every row of Table 1 in turn. And to even number the rows of Table 1 so that you can refer to it better}
\SHREYANK{I've edited the text, if it isn't clear still I'll add the row numbers to the explanation.}


\begin{table}

  \centering
  \resizebox{0.99\linewidth}{!}{ 
  \begin{tabular}{cccccccc}
    \toprule
    
    Index & Spatial Frozen     & Adapter     & Temporal Frozen & Temporal Init & Top-1 Acc & Top-5 Acc & Train Time\\
    \midrule
    - & $\times$ & $\times$ & $\times$ & $\times$ & 64.45 & 87.48 & 14.17 h\\
    I & $\checkmark$ & $\times$ & $\times$ & $\times$ & 27.75 & 56.73 & 0.5x()\\
    II & $\times$ & $\times$ & $\checkmark$ & $\times$ & 25.80 & 53.07 & 0.5x()\\
    III & $\checkmark$ & $\checkmark$ & $\times$ & $\times$ & 38.77 & 68.93 & 0.53x()\\
    IV & $\checkmark$ & $\checkmark$ & $\times$ & $VMAE$ & 58.54 & 85.83 & \textcolor{red}{2.51x()}\\
    V & $\checkmark$ & $\checkmark$ & $\times$ & $ViViT-8f$ & 63.85 & 87.62 & 0.62x()\\

    \bottomrule
  \end{tabular}
  }
    \caption{Ablation study results illustrating the impact of various modifications to the ViViT-B model, including spatial and temporal transformer freezing, adapter addition, and initialization methods, on top-1 and top-5 accuracy. Dataset is Something-something v2.
  \ANURAG{In row 2, the spatial transformer is trained and the temporal transformer is completely randomly initialised?}
  \ANURAG{This table would be more compelling if you also include some computational cost metrics?}\SHREYANK{Added.}
  \JON{can we put captions below the figures/tables?}
  \JON{Also make sure in every table/fig to state what dataset
  the numbers are on}
  \vspace{-6mm}
  }
  \label{tab:ablation}
\end{table}

Additionally, initializing the spatial transformer yields further improvement. The adapter plays a vital role in augmenting performance when the spatial transformer is frozen, and due to its lightweight nature, it will be an essential component of our training methodology moving forward.

\subsection{
How many frames should we use for Stage 1?
}

\begin{figure}[!tb]
    \begin{minipage}{0.45\textwidth}
        \centering
        \includegraphics[width=\linewidth]{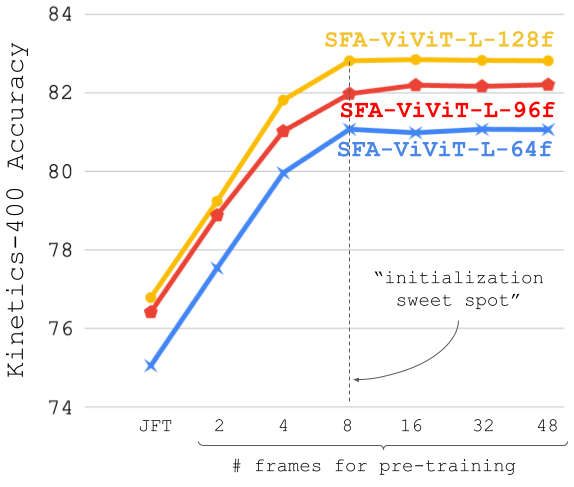}
        \caption{The effect of initializing with different numbers of frames (JFT, 2, 4, 8, 16, 32, and 48), freezing the spatial transformer and adding an adapter model and fine-tuning using 64, 96, and 128 frames. 
        Results on Kinetics400 dataset, `f' refers to frames.}
        \label{fig:diff_frame}
    \end{minipage}%
    \hspace{0.05\textwidth}%
    \begin{minipage}{0.5\textwidth}
        \centering
        \includegraphics[width=\linewidth]{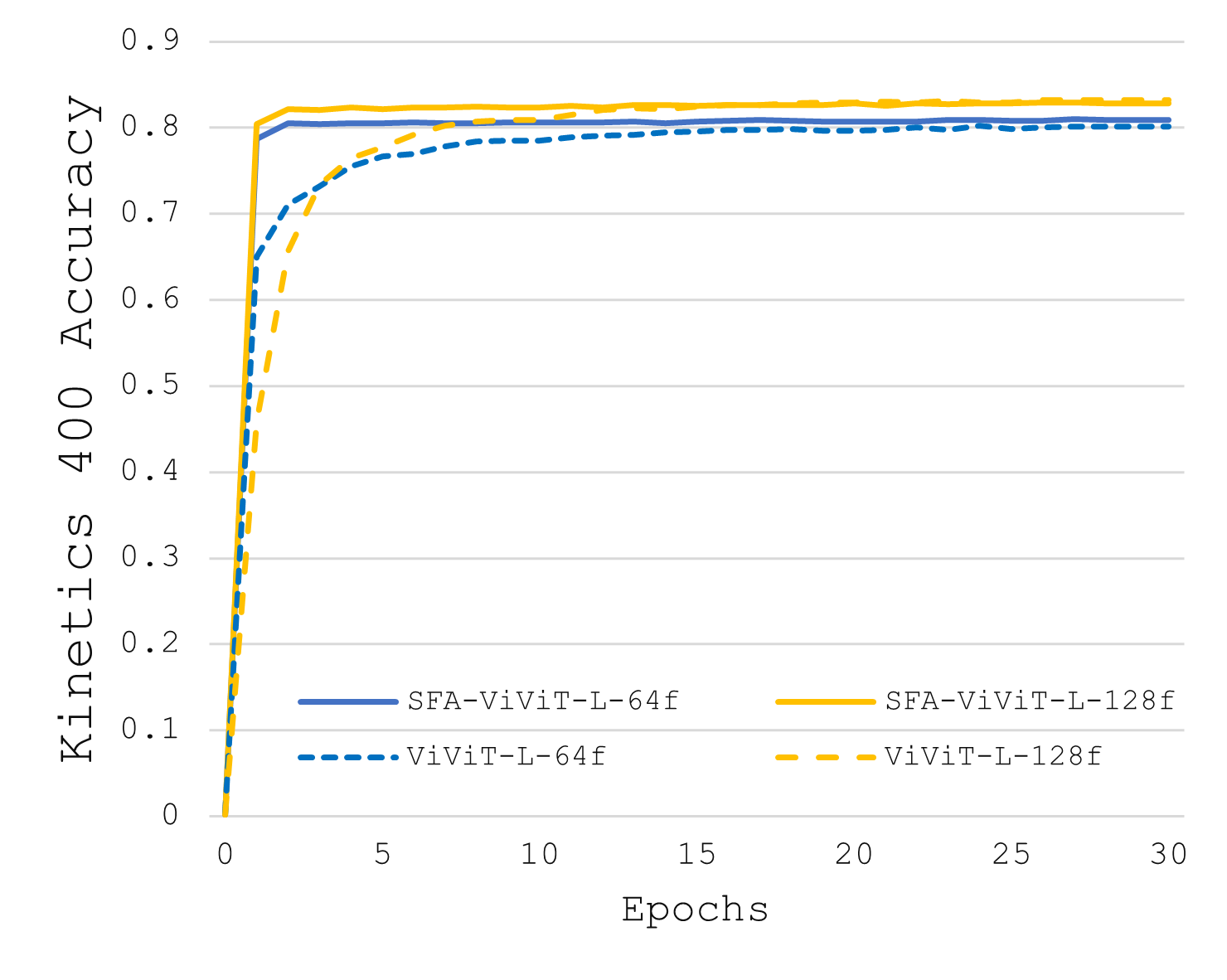}
        \caption{Comparison of our initialization method vs conventional training of ViViT on Top-1 accuracy and loss on the Kinetics-400 dataset using 64 and 128 frames. We see that our initialization gives a significant headstart to the models.}
        \label{plots}
    \end{minipage}
\end{figure}

Next we experiment with various frame counts for stage 1 training,
We test seven  variants: JFT~\cite{jft} checkpoint (image-based), 2, 4, 8, 16, 32, and 48-frame ViViT checkpoints. We then fine-tune these with a frozen spatial transformer and add an adapter model using 64, 96, and 128 frames (see Figure~\ref{fig:diff_frame}). Results show that using 
too few frames for Stage 1 training can underperform 
(with image-only initialization from a JFT~\cite{jft} checkpoint performing the
worst). Thus we deduce that short term temporal context is essential for initializing the spatial transformer.  
Performance also plateaus after 8 frames, and 
given that using more frames increases training time, 
we settle on using 8 frames as our ``sweet spot'' for Stage 1 training. 

\begin{minipage}[t]{0.45\textwidth}
\centering
  \begin{tabular}{ccc}
    \toprule
    
    Checkpoint     & SSv2 & K400 \\
    \midrule
    K400-init & 44.71/74.53 & \textbf{82.81}/\textbf{93.98} \\
    SSv2-init & \textbf{63.85}/\textbf{87.62} & 76.79/92.35 \\
    \bottomrule
  \end{tabular}
\label{tab:crossdataset}
\captionof{table}{A summary of cross-dataset initialization of the proposed model and performance comparison. We use Kinetics400 and Something-something v2 as our datasets.}
\end{minipage}\hspace{0.05\textwidth}%
\begin{minipage}[t]{0.45\textwidth}
\centering\,
  \begin{tabular}{cccc}
    \toprule
    
    Backbone  & Top-1 & Top-5 & Steps \\
    \midrule
ViViT-L &79.64 &91.73 &48k steps \\
ViViT-H &81.02 &93.09 &39k steps \\
ViViT-g & \textbf{81.81} & \textbf{94.55} &29k steps \\

    \bottomrule
  \end{tabular}
\label{tab:vitg}
\captionof{table}{A comparison of top-1 and top-5 accuracies for the ViViT-g model with the proposed training strategy, which incorporates a larger spatial transformer backbone. All models use 48 frames for fair comparison. Results are on Kinetics400 dataset.}
\end{minipage}

\subsection{Does the proposed training increase convergence speed?}

Another potential question that may arise concerns the impact of this initialization method in terms of convergence speed, if any. This specific aspect holds considerable significance owing to its potential ability to drastically curtail the duration of time required for training and the number of epochs necessary to effectively train the model. Moreover, an important element to take into account is the effect of freezing the spatial transformer. This approach decreases the memory needed to store the model but also considerably enhances the training speed. To provide a clearer picture, we have plotted the validation curves with and without initialization, which can be seen in Figure~\ref{plots}. Note that with the proposed initialization, we get a significant head start in overall accuracy.

\subsection{What about initializing on one dataset and finetuning on other?}

In our study, we find training an 8-frame version of the standard ViViT model affordable. We consider training this on a temporally dependent dataset like Something-something, then fine-tuning on other datasets like Kinetics-400. We examine two scenarios: first, the standard ViViT model trained on Kinetics-400 using 8-frames, and second, the same model trained on Something-something. Post-training, we freeze the spatial transformer, add an adapter, and fine-tune the models on the alternate dataset with more frames. We contrast this with models fine-tuned on their original datasets (see Table~\ref{tab:crossdataset}). Results favor initializing larger frame models from the 8-frame version on the same dataset. Thus, for the final comparison in Sec~\ref{benchmark_comparison}, we initialize models with 8-frame versions of the baseline.

\subsection{Extending the image backbone to 1.5B parameters}
\label{sec:vivit_g}

An intriguing consequence of our approach is the ability to incorporate larger backbones into the spatial transformer, made possible by the additional memory available to us as a result of freezing the spatial transformer during training. Consequently, we introduce ViViT-g, which integrates the ViT-g model (with 1.5B parameters)
as its backbone. To ensure a fair comparison, we focus solely on training and inference using 48 frames, and abstain from employing multiview or multicrop \ANURAG{Why?}\SHREYANK{Haha just wanted to keep the difference clearer. I feel with the training of ViT-g on only 4 frames and initializing vs ViT-L and ViT-B on 8 frames results in unfair overall comparison. The overall accuracy using multicrop and multiview results in the same accuracy.}testing. Our objective is to investigate the potential impact of a more substantial spatial transformer backbone on the overall performance and show the potential of larger spatial backbones that are possible due to our training process. 

It is essential to note that the full ViViT-g model could not process more than 8 frames due to memory limitations. However, our proposed strategy allows processing up to 48 frames. A comparison of top-1 and top-5 accuracies is presented in Table~\ref{tab:vitg} along with the number of steps needed to reach the best performance. Dataset used is Kinetics-400 and all the ViT checkpoints are JFT-pretrained~\cite{jft}.

\subsection{Comparison of Memory Usage with Standard ViViT Training and Proposed Method}
\SHREYANK{Added this paragraph cause I think it makes things more impactful with regards to what we are doing. Let me know what you guys think.}
\ANURAG{What you are really talking about here is memory usage. All of this stuff about 64 TPUs or 8x 2080 Ti's or the global batch size is actually irrelevant. It is how much memory training takes on each core. Since the entire model needs to fit on one core.}
\ANURAG{And so, instead of talking about ``Frame accessibility'' just say ``Memory usage'', and how that enables you to process more frames.}
\SHREYANK{Done}
\ANURAG{Rather refer to the local batch size, and you dont need to say the total number of TPUs / GPUs here, since its actually irrelevant}

In this study, we compare the number of frames that can be accessed using the standard ViViT training scheme against our proposed scheme, employing a set of 64 v3 TPUs that have 16 GB each. We further evaluate the performance of ViViT variants, including H, and g, in comparison with the SFA-ViViT using the same variant configurations. Maintaining identical hyperparameters, we ensure a local batch size of 1. 

Our findings indicate that the conventional ViViT training approach restricts frame accessibility to 96 frames for the ViViT-H model, and a mere 8 frames for the ViViT-g model, before reaching memory limitations. 

Conversely, our proposed method enables access to 128 frames for ViViT-H, and up to 48 frames when utilizing ViViT-g with the same hardware. Furthermore, we investigate the impact of utilizing university grade GPUs by conducting ViViT experiments on an NVIDIA Tx 2080 Ti GPU farm equipped with 8 GPUs having 12 GB each. Under these circumstances, ViViT can only process 16 frames using a local batch size of 1. However, our proposed training strategy enables a notable improvement, expanding the frame capacity to 80 frames helping us reproduce ViViT results on lower end GPUs. This enhancement provides a valuable opportunity for researchers with limited resources to attain performance levels comparable to those with extensive resources. We show a comparison of number of frames accessible with and without our training recipe in Table~\ref{tab:frame_access}.

\begin{table}

  \centering
  \resizebox{0.6\linewidth}{!}{ 
  \begin{tabular}{ccccccccccc}
    \toprule
    
    Model & 4 & 8 & 16 & 32 & 48 & 64 & 96 & 128\\
    \midrule
    ViViT-H & $\checkmark$ & $\checkmark$ & $\checkmark$ & $\checkmark$ & $\checkmark$ & $\checkmark$ & $\checkmark$ & $\times$\\
    SFA-ViViT-H & $\checkmark$ & $\checkmark$ & $\checkmark$ & $\checkmark$ & $\checkmark$ & $\checkmark$ & $\checkmark$ & $\checkmark$\\
    ViViT-g & $\checkmark$ & $\checkmark$ & $\times$ & $\times$  & $\times$  & $\times$  & $\times$ & $\times$ \\
    SFA-ViViT-g & $\checkmark$ & $\checkmark$ & $\checkmark$ & $\checkmark$ &$\checkmark$ & $\times$  & $\times$  & $\times$ \\

    \bottomrule
  \end{tabular}
  }
  \caption{Memory usage in different ViViT training schemes is compared using the Kinetics400 dataset on 64 TPUs v3 with 16GB memory each. A $\checkmark$ indicates accessible frames given hardware constraints, while a $\times$ signals an out-of-memory (OOM) error.
  \ANURAG{You need to say how much accelerator memory you have? I guess you used PF-Megacore? (which is 32GB then). Wait, looks like you used DF then, that's 16 GB.}\SHREYANK{I've added this in the text. Do I need to change in table too?}
  }
  \label{tab:frame_access}
\end{table}

\subsection{Comparison on all benchmarks to the baseline model}
\label{benchmark_comparison}

In this section, we present a comprehensive comparative analysis, focusing on the proposed approach and the baseline model. We report the Top-1 accuracy, Top-5 accuracy and the overall training time.

The evaluation is conducted on the large and huge variants of ViViT across three datasets, namely Kinetics400, Kinetics600, and Moments in Time (MiT), with the summarized results tabulated in Table~\ref{tab:final_1}. The findings indicate a slight enhancement in accuracy for both Kinetics400 and Kinetics600 datasets, whereas a notable 1.79\% increase in top-1 accuracy is observed for the MiT dataset using the proposed method. 

Furthermore, the proposed approach showcases a significant reduction in training time, accounting for approximately 56\% of the original duration. This reduction emphasizes the advantageous nature of the proposed approach. To calculate the total training time for the SFA version, the train time of the 8 frame (Stage 1) ViViT model is combined with the train time of the (Stage 2) SFA-ViViT model. Conversely, the total training time for the standard ViViT encompasses the total train time for the same number of frames that SFA-ViViT is trained on for fair comparison.
\ANURAG{How do you calculate the training time? Should show somewhere the time taken for each step of training.}
\SHREYANK{I was adding this but then realized that this is kind of a curriculum training mechanism and was not sure how to add time taken for each step given we also include training the 8f ViViT time to the train time in the table. What should I do? I've written a line about it.}
\ANURAG{I don't understand by ``the total training time for the standard ViViT encompasses the cumulative time required to complete all training steps''? Standard ViViT has one training stage. So what is ``cumulative'' here?}
\SHREYANK{Sorry fixed now!}

\begin{table}
  \centering
  \resizebox{0.99\linewidth}{!}{ 
  \begin{tabular}{lllllllll}
    \toprule
    \multicolumn{1}{c}{Model} & \multicolumn{2}{c}{Kinetics-400} & \multicolumn{2}{c}{Kinetics-600} & \multicolumn{2}{c}{Moments in Time} \\
    \multicolumn{1}{c}{} & \multicolumn{1}{c}{Accuracy}  & \multicolumn{1}{c}{Train Time} & \multicolumn{1}{c}{Accuracy} & \multicolumn{1}{c}{Train Time} & \multicolumn{1}{c}{Accuracy}  & \multicolumn{1}{c}{Train Time}      \\
    \midrule
    ViViT-L & 82.59/93.09  & 1x(21.57 h) & 83.29/\textbf{95.82} & 1x(26.14 h) & -& - & -\\
    ViViT-L + \textbf{SFA}  & \textbf{82.78/94.03}  &\textbf{0.56x()} & \textbf{83.47}/95.29 & \textbf{0.56x()} & - & - & \\
    ViViT-H & 84.21/94.66  & 1x(56.71 h) & 84.18/95.68 & 1x(60.45 h)  & 38.17/62.84  & 1x(110.79 h) \\
    ViViT-H + \textbf{SFA} & \textbf{84.42/94.72}  & \textbf{0.57x()} & \textbf{84.39/96.20}& \textbf{0.57x()}  & \textbf{39.96/64.39}   & \textbf{0.59x()}  \\
    \bottomrule
  \end{tabular}
  } 
  \caption{Performance Comparison of various versions of ViViT with the proposed training strategy for Kinetics-400, Kinetics-600 and Moments in Time. Accuracies listed as Top-1/Top-5.}
\label{tab:final_1}

  \centering
  \resizebox{0.99\linewidth}{!}{ 
  \begin{tabular}{lllllll}
    \toprule
    \multicolumn{1}{c}{Model} & \multicolumn{2}{c}{Something-something} & \multicolumn{2}{c}{Something-Else} & \multicolumn{2}{c}{Epic-Kitchens} \\
    \multicolumn{1}{c}{} & \multicolumn{1}{c}{Accuracy} & \multicolumn{1}{c}{Train Time} & \multicolumn{1}{c}{Accuracy} & \multicolumn{1}{c}{Train Time} & \multicolumn{1}{c}{Accuracy} & Train Time       \\
    \midrule
    ViViT-L & \textbf{64.45}/87.48 & 1x(14.17 h) & 53.14/73.98 & 1x(3.84 h) & 43.53/56.55/\textbf{65.40} & 1x(5.61 h)\\
    ViViT-L + \textbf{SFA}  & 63.85/\textbf{87.62} & \textbf{0.62x()} & \textbf{53.60/74.47} & \textbf{0.62x()} & \textbf{43.54/56.78}/65.16 & \textbf{0.63x()} \\
    \bottomrule
  \end{tabular}
  } 
  \caption{Performance Comparison of ViViT-L with the proposed training strategy for Something-something v2, Something-Else and Epic-Kitchens. Accuracies listed as Top-1/Top-5, for Epic Kitchens Top-1 noun-verb/ Top-1 noun/ Top-1 Verb.}
  \label{tab:final}
\end{table}

We further examine the performance of ViViT-L incorporating our proposed training strategy in comparison to the original version on three additional datasets: Something-something, Something-Else, and Epic-Kitchens. A consistent trend is observed, with the modified approach outperforming the baseline model, 
at only a 62\% cost of the baseline training time. In summary, our proposed training strategy demonstrates promising potential by yielding comparable or slightly improved performance across all datasets. This is obtained while maintaining a training cost ranging from 56\% to 62\% of the original model, thus highlighting its effectiveness. Results can be seen in Table~\ref{tab:final}.

\section{Limitations}
\label{limitations}

Our research makes considerable progress in reducing training time and memory use for video transformers, but it raises certain issues. First, training smaller versions of our model on different datasets is required, adding an initial step. Ideally, a universal model applicable across datasets would improve efficiency. Our method depends on separate space and time encoders, a feature of the ViViT model, which might limit its use with integrated space-time models. We base our work on the ViViT model used in influential models like MTV, highlighting its importance. While we didn't test our methods on models like MTV, focusing on ViViT provides beneficial implications for other models. We hope this inspires future research and encourages further exploration in efficient training of video transformers. \SHREYANK{Need to make sure that the weaknesses are not exploited for reviews intending to reject the paper} 
\ANURAG{You can shorten this paragraph a lot}\SHREYANK{Shortened now.}

\section{Conclusion}
\label{conclusion}

We have investigated the challenges posed by the substantial training time and memory consumption of video transformers, particularly focusing on the factorised encoder variant of the ViViT model as our baseline. To address these challenges, we proposed two effective strategies: utilizing a compact adapter model for fine-tuning image representations instead of end-to-end training of the spatial transformer, and initializing the temporal transformer using the baseline model trained with 8 frames. Our proposed training strategy has demonstrated the potential to significantly reduce training costs and memory consumption while maintaining, or even slightly improving, performance compared to the baseline model. Furthermore, we observed that with proper initialization, our baseline model can achieve near-peak performance within the first 10\% of training epochs. The advancements made in this work have the potential to propel research in the video understanding domain by enabling access to more frames and the utilization of larger image models as the spatial transformer, all while maintaining the same memory consumption. Our findings provide valuable insights for researchers and practitioners with limited resources, paving the way for more efficient and scalable alternatives in the action recognition field. Future work may focus on further optimizing and refining these strategies, and exploring their application to other video transformer architectures and tasks in the computer vision domain.

\appendix

\section{ViViT hyperparameters}

\begin{table}[h]
    \centering
\begin{tabular}{|c|c|c|c|c|c|c|}
\hline & \textbf{K400} & \textbf{K600} & \textbf{MIT} & \textbf{Epic-Kitchens} & \textbf{SSv2} & \textbf{Selse}\\
\hline \multicolumn{7}{|l|}{ \textbf{Optimisation} } \\
\hline Optimiser & \multicolumn{6}{|c|}{ Synchronous SGD } \\
\hline Momentum & \multicolumn{6}{|c|}{0.9} \\
\hline Batch size & \multicolumn{6}{|c|}{128} \\
\hline Learning rate schedule & \multicolumn{6}{|c|}{ cosine with linear warmup } \\
\hline Linear warmup epochs & \multicolumn{6}{|c|}{2.5} \\
\hline Base learning rate & 0.1 & 0.1 & 0.25 & 0.5 & 0.5 & 0.5 \\
\hline Epochs & 30 & 30 & 10 & 50 & 35 & 35 \\
\hline \multicolumn{7}{|l|}{ \textbf{Data augmentation} } \\
\hline Random crop probability & \multicolumn{6}{|c|}{1.0} \\
\hline Random flip probability & \multicolumn{6}{|c|}{0.5} \\
\hline Scale jitter probability & \multicolumn{6}{|c|}{1.0} \\
\hline Maximum scale & \multicolumn{6}{|c|}{1.33} \\
\hline Minimum scale & \multicolumn{6}{|c|}{0.9} \\
\hline Colour jitter probability & 0.8 & 0.8 & 0.8 & - & - & -\\
\hline Rand augment number of layers \cite{cubuk2020randaugment} & - & - & - & 2 & 2 & - \\
\hline Rand augment magnitude \cite{cubuk2020randaugment} & - & - & - & 15 & 20 & - \\
\hline \multicolumn{7}{|l|}{ Other regularisation } \\
\hline Stochastic droplayer rate, $p_{\text {drop }}$ \cite{pdrop} & - & - & - & 0.2 & 0.3 & - \\
\hline Label smoothing \cite{szegedy2016rethinking} & - & - & - & 0.2 & 0.3 & -\\
\hline Mixup \cite{mixup} & - & - & - & 0.1 & 0.3 & -\\
\hline
\end{tabular}
    \caption{The hyperparameters utilized in the experiments conducted for the primary research paper are detailed here. If a regularisation method is not employed, it is represented by a "–". Constant values that are present across all columns are mentioned just once. For simplicity, abbreviations have been used to denote different datasets: Kinetics 400 is represented as K400, Kinetics 600 as K600, Moments in Time as MiT, Epic Kitchens as EK, Something-Something v2 as SSv2 and Something-Else as Selse.}
    \label{tab:hyperparams}
\end{table}

 We have already mentioned the hyperparameters for the various transformer sizes used. In Table~\ref{tab:hyperparams} we list the hyperparameters used for each dataset. For fair comparison we re-run SFA-ViViT using the same hyperparameters as ViViT.

\section{Datasets}

As Kinetics consists of YouTube videos which may be removed by their original creators, we note the exact sizes of our dataset.

Kinetics-400 \cite{i3d}: 
Kinetics-400 is a large-scale video dataset with 400 classes introduced by Google's DeepMind. It has 235693 training samples and 53744 validation and test samples. The dataset encompasses various categories, such as object manipulation, human-object interaction, and body movements. Each class contains approximately 400 video samples, with each video lasting around 10 seconds.

Kinetics-600 \cite{k600}:
Kinetics-600 is an extension of the Kinetics-400 dataset, with an increased number of classes, totaling 600 human action classes. This dataset contains approximately 380735 training samples and 56192 validation and test samples. The additional classes broaden the scope of the dataset, thereby providing more diverse training data for video recognition tasks.

EPIC Kitchens \cite{epic}:\SHREYANK{Need to check details about this as internal feedback suggested this is older version. Need help to finetune numbers on all datasets.}
EPIC Kitchens is a large-scale dataset focusing on egocentric (first-person) videos of daily kitchen activities. It consists of 55 hours of video captured by 32 different participants in their own kitchens, with 67217 training samples and 22758 samples for validation and testing. The dataset includes 97 verb classes and 300 noun classes. Epic Kitchens is particularly useful for understanding human-object interactions and fine-grained actions in everyday settings.

Something-something v2 \cite{ssv2}:
The Something-something v2 dataset is a collection of short video clips focused on common objects and human actions. It contains around 168913 training clips and 24777 test clips distributed across 174 action classes. This dataset aims to capture more abstract and high-level understanding of actions, as well as temporal relationships among objects.

Moments in Time \cite{mit}:
The Moments in Time dataset is a large-scale video dataset containing one million short video clips, each lasting three seconds. It covers 339 classes of dynamic events and aims to provide a diverse set of visual and auditory representations of these events with 791297 training samples and 33900 test samples. This dataset is particularly useful for understanding the temporal aspects of various activities and events, as well as their associated contexts.

Something-else \cite{selse}:
Something-Else utilizes the videos from SomethingSomething-V2 as its foundation, and introduces novel training and testing partitions for two new tasks that examine the ability to generalize: compositional action recognition and few-shot action recognition. Our attention is solely on the compositional action recognition task, which aims to prevent any object category overlap between the 54919 training videos and the 57876 validation videos.

\section{How important is the pre-training image dataset for action recognition performance?}

While we know from the original ViViT paper ~\cite{vivit} that using larger ViT~\cite{vit} backbones result in better performances, we do a more thorough ablation here by considering variations of the ViT model such as the hybrid ViT (ResNet-ViT-L pre-trained on ImageNet21k ~\cite{ridnik2021imagenet}), ViT-L pre-trained on ImageNet21k, ViT-L pre-trained on JFT and ViT-H pre-trained on JFT. We report these results in Table~\ref{tab:pretrained}, with the  conclusion of larger backbones pre-trained on larger datasets yields highest accuracies. We report top-1 and top-5 accuracies on the Kinetics-400 dataset and we freeze the spatial transformer here without any fine-tuning or adapter. We also keep the temporal transformer fixed in size here for fair comparison. Essentially, the performance difference is purely from the output of the spatial transformer changing due to different backbones.
\ANURAG{So what is the main point of this? What is it showing that is not in the ViViT paper? (ablations of backbone size and number of frames are both in the paper). And more so, what is the relevance of this?}
\SHREYANK{I could remove this cause these are mainly results of having a fully frozen spatial transformer and the impact of larger backbones and bigger pre-training datasets.}

\begin{table}

  \centering
  \begin{tabular}{cccc}
    \toprule
    
    Backbone  & 16-frames & 32-frames & 48-frames \\
    \midrule
    ResNet-ViT-L (ImageNet21k) & 66.09/88.30 & 66.63/88.50 & 66.88/88.65 \\
    ViT-L (ImageNet21k) & 65.59/85.86 & 68.34/87.80 & 70.09/88.91 \\
    ViT-L (JFT) & 69.76/88.41 & 73.98/90.88 & 75.08/91.69 \\
    ViT-H (JFT) & 73.68/90.23 & 75.85/91.53 & 77.90/92.72 \\
    \bottomrule
  \end{tabular}
  
  \caption{Comparison of impact of different backbones for the spatial transformer. We use ResNet-ViT-L pre-trained on ImageNet21k, ViT-L pre-trained on ImageNet21k, ViT-L pre-trained on JFT and ViT-H pre-trained on JFT. Listed as (Top-1 accuracy/ Top-5 accuracy).}
  \label{tab:pretrained}
\end{table}

\section{Curriculum Training}

\begin{figure}
  \centering
  \includegraphics[width=0.9\linewidth]{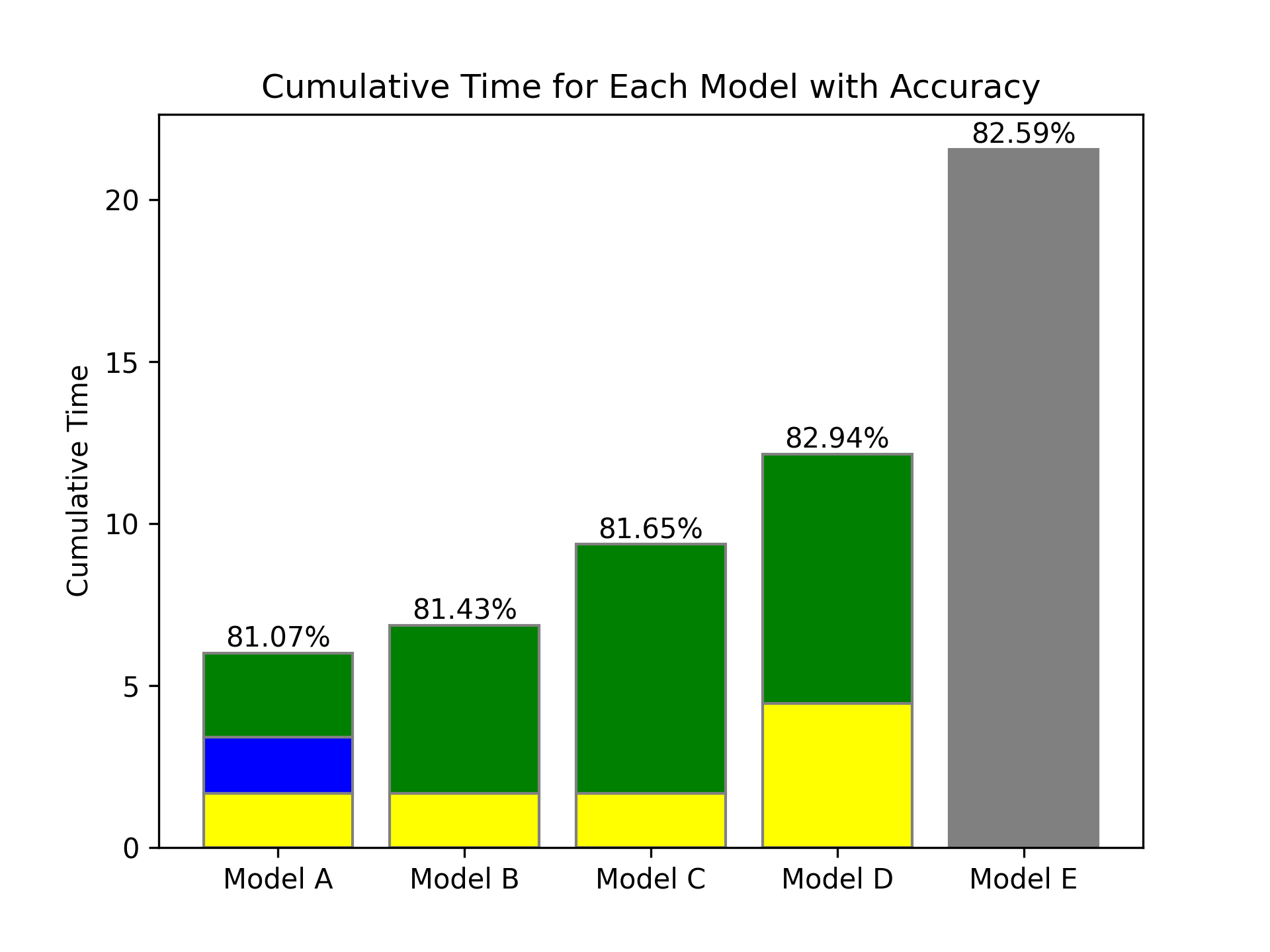}
  \caption{Stacked bar chart representing the cumulative processing times of Models A-E \JON{what are models
  A-E?}. Each color within a bar corresponds to a specific sub-model ('a' in yellow, 'b' in blue, 'c' in green, 'd' in gray) contributing to the total computation time of each model. Model accuracies are indicated at the top of each respective bar. 'a' = ViViT-L-8f, 'b' = SFA-ViViT-L-32f, 'c' = SFA-ViViT-L-128f, 'd' = ViViT-L-128f. All results are using Kinetics-400 dataset and using ViViT-L variants.
  }
  \label{fig:curriculum}
\end{figure}

We consider variants of the ``curriculum" training we talk about in the paper. There are various forms that we can consider. For instance, we can train the standard ViViT 8 frame model for just 10 epochs and use that to initialize our model. In the paper all initializations are done using 8 frame model trained for 30 epochs. Further, we could initialize smaller versions of SFA-ViViT like a 32 frame version for 10 epochs and then initialize SFA-ViViT 128 frames using this 32 frame version. We plot this in Figure~\ref{fig:curriculum} and see various versions and conclude that in the end, the best speed-accuracy trade-off was obtained when the standard ViViT 8 frame model was trained on 30 epochs and then the 128 frame model is initialized using this.

We define the models in the figure as follows:
\begin{itemize}
    \item Model A: ViViT-L-8f for 10 epochs + SFA-ViViT-L-32f for 10 epochs + SFA-ViViT-L-128f for 10 epochs
    \item Model B: ViViT-L-8f for 10 epochs + SFA-ViViT-L-128f for 20 epochs
    \item Model C: ViViT-L-8f for 10 epochs + SFA-ViViT-L-128f for 30 epochs
    \item Model D: ViViT-L-8f for 30 epochs + SFA-ViViT-L-128f for 30 epochs
    \item Model E: ViViT-L-128f for 30 epochs
\end{itemize}

We see that training the ViViT-L-8f model for the full 30 epochs and then using that to initialize the SFA-ViViT-L-128f model gave us the best results. But we could potentially reduce the cost of training to 0.25x if we sacrifice 2\% accuracy. All results are on the Kinetics400 dataset.

\section{How long do we need to train the model?}

We showed in the paper that using SFA based initialization helps us reach ``near-peak'' performance really quickly. We define this near-peak performance as 1 \% less than the eventual best performance of the model. 
Thus another natural question is:
in order to save time, why not stop training the SFA version earlier? We note that although the standard ViViT model trains for `x' epochs (see Table.~\ref{tab:hyperparams} for exact number), it often reaches this ``peak'' performance much earlier and hence for fair comparison with the standard ViViT model, in the paper, we run on the same number of epochs. These results can be seen in Table~\ref{tab:pretrained_}. 
\JON{point to table.}

\begin{table}

  \centering
  \begin{tabular}{cccc}
    \toprule
    
    Model  & Dataset & NPP Epoch & Best Epoch \\
    \midrule
    ViViT-L & K400 & 20 & 29\\
    SFA-ViViT-L & K400 & 5 & 28\\
    ViViT-L & K600 & 21 & 28 \\
    SFA-ViViT-L & K600 & 5 & 23 \\
    ViViT-L & SSv2 & 29 & 35\\
    SFA-ViViT-L & SSv2 & 4 & 24\\
    \bottomrule
  \end{tabular}
  
  \caption{Comparison of near peak performance (NPP) epoch and best performance epoch for ViViT and SFA-ViViT for different datasets and models. All results are on Kinetics400 dataset.
  \JON{caption is identical to one below and labels
  are identical too!}
  }
  \label{tab:pretrained_}
\end{table}

\section{What about initializing standard ViViT models?}

Since our method proposes an initialization scheme, we also test it on the standard ViViT models that do not have their spatial transformer frozen. In this particular scenario, we only want to check if the peak performance can be reached faster. However, it is important to note that with our proposed training scheme we also reduce the overall training time by close to half. This can be seen in Table~\ref{tab:init}.
\JON{point to table.}

\begin{table}

  \centering
  \begin{tabular}{cccc}
    \toprule
    
    Model  & Dataset & NPP Epoch & Best Epoch \\
    \midrule
    ViViT-L & K400 & 20 & 29\\
    ViViT-L init with 8f ViViT-L & K400 & 4 & 25\\
    ViViT-H & K400 & 22 & 27 \\
    ViViT-H init with 8f ViViT-H & K400 & 5 & 22 \\
    \bottomrule
  \end{tabular}
  
  \caption{Comparison of near peak performance (NPP) epoch and best performance epoch for initializing the full ViViT model with and without the 8f variant. We see the benefit of initialization as the ``near-peak" performance is reached at a much earlier stage when initialized with the 8f variant. All results are on Kinetics400 dataset.}
  \label{tab:init}
\end{table}

{\small
\bibliographystyle{abbrv}
\bibliography{sfa}
}

\end{document}